\documentclass[10pt,twocolumn,letterpaper]{article}

\usepackage{iccv}
\usepackage{times}
\usepackage{epsfig}
\usepackage{graphicx}
\usepackage{amsmath}
\usepackage{amssymb}

\usepackage{bm}
\usepackage{wrapfig}
\usepackage{tikz}
\usepackage{scalerel}
\usepackage{xspace}
\usepackage{pifont}
\usepackage{multirow}
\usepackage{tikz}
\usepackage{comment}
\usepackage{amsmath,amssymb} 
\usepackage{color}
\usepackage[inline]{enumitem}
\usepackage{float}
\usepackage{booktabs}       
\usepackage{indentfirst}
\usepackage{multirow}
\usepackage{makecell}
\usepackage{pifont}
\usepackage{everyshi}

\newcommand{\cmark}{\ding{51}}%
\newcommand{\xmark}{\ding{55}}%
\usepackage{caption} 
\usepackage[breaklinks=true,bookmarks=false]{hyperref}

\iccvfinalcopy 

\def\iccvPaperID{2932} 

\ificcvfinal\fi

\begin{document}

\title{Hallucination Improves the Performance of Unsupervised Visual Representation Learning}

\author{Jing Wu$^{1}$ $\quad$ Jennifer Hobbs$^{2}$ $\quad$ Naira Hovakimyan$^{1}$\\
$^{1}$University of Illinois Urbana-Champaign, $\qquad$$^{2}$Intelinair\\
{\tt\small  $\left\{jingwu6, nhovakim\right\}@illinois, \qquad  jenniferhobbs08$+$research@gmail.com$}
\and
}

\maketitle
\ificcvfinal\thispagestyle{empty}\fi

\begin{abstract}
Contrastive learning models based on Siamese structure have demonstrated remarkable performance in self-supervised learning. Such a success of contrastive learning relies on two conditions,  a sufficient number of positive pairs and adequate variations between them. If the conditions are not met, these frameworks will lack semantic contrast and be fragile on overfitting. To address these two issues, we propose \textit{Hallucinator} that could efficiently generate additional positive samples for further contrast. The Hallucinator is differentiable and creates new data in the feature space. Thus, it is optimized directly with the pre-training task and introduces nearly negligible computation. Moreover, we reduce the mutual information of hallucinated pairs and smooth them through non-linear operations. This process helps avoid over-confident contrastive learning models during the training and achieves more transformation-invariant feature embeddings. Remarkably, we empirically prove that the proposed Hallucinator generalizes well to various contrastive learning models, including MoCoV1\&V2, SimCLR and SimSiam. Under the linear classification protocol, a stable accuracy gain is achieved, ranging from  $0.3\%$ to $3.0\%$ on CIFAR10\&100, Tiny ImageNet, STL-10 and ImageNet. The improvement is also observed in transferring pre-train encoders to the downstream tasks, including object detection and segmentation.
\end{abstract}

\section{Introduction}
In the recent computer vision community, there has been rapid progress in self-supervised learning (SSL), gradually closing the performance gap with supervised learning \cite{grill2020bootstrap,he2020momentum,chen2020simple,caron2020unsupervised,tian2020makes}. Among the diverse approaches of SSL, contrastive learning, such as MoCoV1\&V2 \cite{he2020momentum,chen2020improved}, SimCLR \cite{chen2020simple}, and SimSiam \cite{chen2021exploring}, shows promising results. Generally, contrastive learning treats each image as one class which will be augmented into two separate views. These two views form one positive pair and should ideally be close if mapped to feature space. With sufficient contrast in the feature space, contrastive learning models show a strong capacity to learn transformation-invariant features that are transferable to various downstream tasks, such as classification, object detection, and segmentation  \cite{everingham2009pascal,he2017mask, lin2014microsoft}.

\begin{figure}[t]
\begin{center}
    \includegraphics[width=0.9\linewidth]{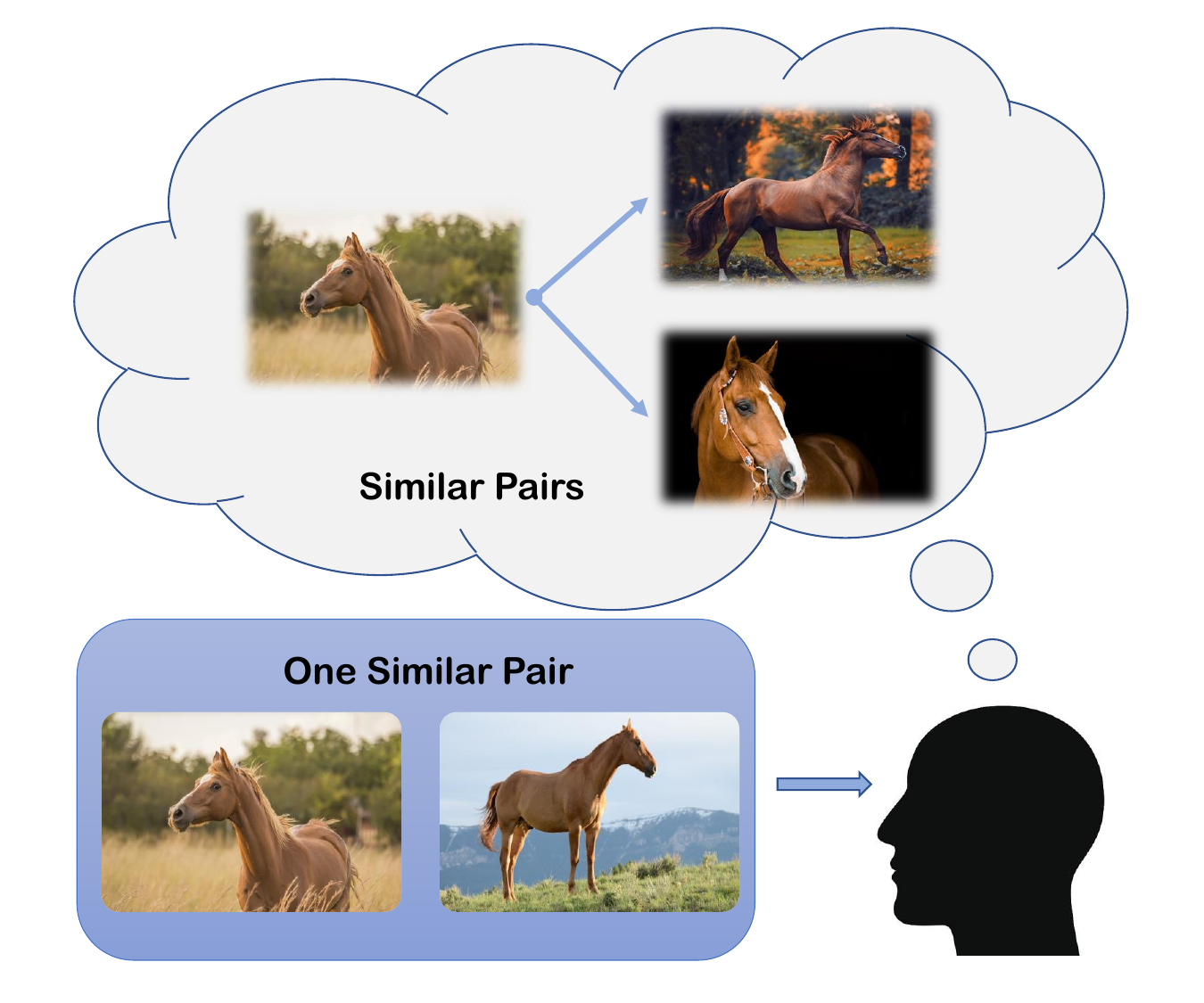}
\end{center}
  \vspace*{-3mm} 
  \caption{The motivation of the proposed hallucination methods. Given one pair of images with the same semantic meaning, such as a pair of horses, a person can envision further similar pairs by imagining one of the horses in different poses and surroundings. If a contrastive learning model could do such hallucination, it could have additional novel pairs to contrast given the same data. Note that this hallucination process is for illustration only. In the implementation, all the hallucinated samples are computed in the feature space.}
  \vspace*{-3mm} 
\label{fig: demo}
\end{figure}
To ensure sufficient contrast, researchers from previous work address the issue from two essential practices, either introducing large amounts of positive pairs or adding additional variants\&transformation among them. For example, SimCLR uses a batch size that generates thousands of positives to facilitate the convergence of models \cite{chen2020simple}. Work from \cite{tian2020makes, chen2021jigsaw} reduces mutual information of positive pairs using stronger data augmentation, i.e., color distortion and jigsaw transformation. Likewise, the work from \cite{peng2022crafting} introduces ContrastiveCrop, and the work from \cite{shen2022mix} proposes Un-Mix, respectively, to reduce the similar semantic meaning of sample pairs in the original image space. Beyond the data augmentation and image operations, researchers from \cite{zhu2021improving} propose to apply a linear operation to generate hard positive samples in feature space.

Despite the success of prior approaches,  we argue that large batch sizes are not always achievable. Meanwhile, all proposed techniques only focus on improving the original pairs. Given one positive pair of positive samples, humans are born with the amazing ability to come up with additional positives by imagining a sample from different surroundings and perspectives without much effort, as demonstrated in Figure~\ref{fig: demo}. This process of self-imagination, in turn, will benefit the human neurological system, improving recognition capacity \cite{grilli2011self}. Similarly, if we could empower contrastive learning models with the ability to hallucinate or imagine an object to a novel view, additional positive pairs could be provided for the learning tasks.

Unfortunately, exploring feasible methods to hallucinate novel positive pairs is challenging. Firstly, while generative models produce realistic images that could form additional positive views  \cite{goodfellow2020generative,arjovsky2017wasserstein,miyato2018spectral,brock2018large}, realistic data do not necessarily benefit learning tasks \cite{wang2018low}. More importantly, applying these approaches forces us to fall back into a computational dilemma to the previous method. In other words, image-level hallucination still suffers from expensive computation as we still need to encode the hallucinated images into feature space. Lastly, if the generated positive pairs are similar to each other, training a discriminative model would be too trivial, thus showing poor generalization capacity \cite{peng2022crafting,tian2020makes,zhu2021improving}.

Therefore, our key insight is that the sample-generation process should aim for three critical elements: (i) feature-space operation (ii) sufficient variance of positive pairs (iii) a differentiable module optimized directly related to the learning task. To achieve this, we propose \textit{Hallucinator} to improve the performance of contrastive learning with Siamese structures. The \textit{Hallucinator} is plugged in after the encoder to manipulate feature vectors and improve the feature-level batch size for further contrast. To ensure adequate variance is introduced, we propose an asymmetric feature extrapolation method inspired by the work from \cite{zhu2021improving}. More importantly, we present a non-linear hallucination process for the extrapolated samples. Such a process is differentiable (i.e. learnable), therefore essentially boosting \textit{Hallucinator} to generate smooth and task-related features.

The proposed \textit{Hallucinator} delivers extra positives and simultaneously enlarges the variance between newly introduced pairs. Moreover, this approach only relies on positive samples. Therefore, it can be easily applied to any Siamese structure by adding it after the encoders as a plug-and-play module. Without the tedious exploration of hyper-parameters and much additional computation, we empirically prove the effectiveness of the proposed  \textit{Hallucinator} on popular contrastive learning models, including MoCoV1\&V2, SimCLR and SimSiam. We notice a stable improvement ranging from $0.3\%$ to $3.0\%$ under the linear classification protocol, crossing the CIFAR10\&100, Tiny ImageNet, STL-10 and ImageNet. We also observe that models trained with \textit{Hallucinator} show better transferability in downstream tasks like object detection and segmentation.

Our contributions can be summarized as follows:
\begin{itemize}[noitemsep,topsep=0pt]

    \item[$\bullet$] We investigate a critical yet under-explored aspect of contrastive learning:  introducing additional positive pairs with further variation in feature space. 

    \item[$\bullet$]To the best of our knowledge, this is the first attempt to incorporate the concept of ``Hallucination" into contrastive representation learning. We propose \textit{Hallucinator} to realize this idea, which effectively generates smooth and less similar positive feature vectors. The \textit{Hallucinator} is simple, effective, and agnostic to contrastive frameworks.

    \item[$\bullet$]We empirically illustrate that the proposed approach significantly benefits the various contrastive learning models within multiple datasets.

\end{itemize}

\begin{figure*}[t!]
\begin{center}
    \includegraphics[width=1\linewidth]{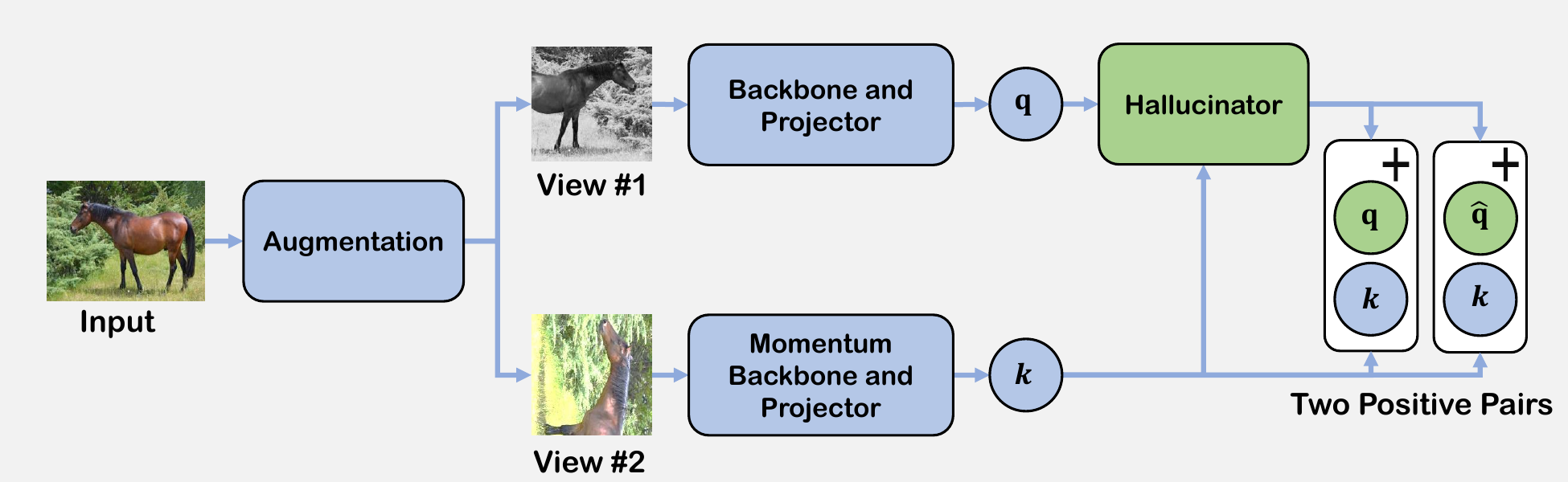}
\end{center}
  \caption{Illustration of contrastive learning (MoCoV2 \cite{chen2020improved}) with $\textit{Hallucinator}$. The $\textit{Hallucinator}$ is added after the backbones and projector for feature-level manipulation. While $\textit{Hallucinator}$ feeds original feature vector $q$ forward, it additionally provides hallucinated feature $\hat{q}$ for further contrast. }
\label{fig: pipline}
\end{figure*}

\section{Related Works}
In this section, we introduce related literature on contrastive learning and hallucination techniques.

\subsection{Contrastive Learning}
The key idea of contrastive learning is to minimize the distance of positive pairs and repulse negative pairs in the feature space \cite{hadsell2006dimensionality}. This idea has been successfully applied to unsupervised visual representation tasks, showing promising results in various downstream tasks  \cite{chen2020simple,he2020momentum,chen2020improved,grill2020bootstrap,chen2021exploring,bachman2019learning,henaff2020data,dwibedi2021little,wang2021dense,misra2020self,oord2018representation,tian2020contrastive,wu2023extended,wu2018unsupervised,ye2019unsupervised,xie2021propagate}. One of the breakthroughs of contrastive learning models is SimCLR   \cite{chen2020simple}, which introduces a simple but effective visual representation learning method. Without negative pairs, SimCLR learned transformation-invariant representation with a large batch size. Contemporaneous impressive work is MoCo from \cite{he2020momentum}. To ensure MoCo can be trained smoothly with computational-friendly batch sizes, the authors of MoCo propose a memory bank to store negative features and momentum-updated backbones. After this, SimSiam \cite{chen2021exploring} presents a Siamese-based network that can learn high-quality representation with stop-gradient, successfully avoiding the collapse of contrastive learning models. Other designs of contrastive learning rely on an online network to predict the output of the target network or contrasting cluster assignments \cite{grill2020bootstrap,caron2020unsupervised}. 

To further boost the models' performance, researchers explore diverse ways to reduce the mutual information of positive samples so that more transformation-invariant embeddings can be learned \cite{tian2020makes}. For image-level enhancement, typical works focus on improving the data augmentation in pixel space using Mixup, color distortion, or jigsaw transformation \cite{shen2022mix, tian2020makes, chen2021jigsaw}. More recent work is presented by \cite{peng2022crafting} using ContrastiveCrop, which creates better positive views through the localization box. For feature-level operation, the authors from \cite{kalantidis2020hard} prove the effectiveness of hard negative samples. More recent and close work to this paper is  \cite{zhu2021improving}, which applies symmetric extrapolation to create hard positives but introduces no non-linearity. Importantly, instead of replacing the original positive samples, this paper proposes a differentiable hallucinator to generate an additional positive sample with less mutual information and better smoothness. Such a setting ensures maximal and adaptive contrast during the training, generalizing well to diverse contrastive learning models.

\subsection{Hallucination}

Hallucination is initially proposed to solve the scarcity of data in the classification task \cite{hariharan2017low}. Then, this idea is kept updated and applied in different areas \cite{schwartz2018delta,wang2018low,zhang2018metagan,zhang2021hallucination,cao2017attention,zhang2019few,gui2021learning,shah2023halp}, such as object detection, aerial navigation, skeleton-based action recognition, and face generation. While image-level hallucination benefits few-shot recognition by synthesis of novel view \cite{bao2020bowtie} or introducing random noises \cite{wang2018low}, most of the work applies hallucination in the feature space. The work from \cite{hariharan2017low} generates novel class features by transforming shared features in base classes. Authors of \cite{zhang2021hallucination} build a hallucination framework in the region of interest feature space object to enhance object detection performance. More recent work shows that this hallucination mechanism also benefits 3D human pose estimation by generating novel motion sequences \cite{gong2022posetriplet}. While hallucination is effective in different learning tasks, to the best of our knowledge, the performance and application of hallucination in SSL are fully unexplored. 

\subsection{Feature-Level Augmentation}

Hallucination relies on effective feature-level augmentation or manipulations. The primary goal of feature augmentation is to extend the limited labeled dataset without relying on expensive computation, such as Generative Adversarial Networks\cite{mariani2018bagan} or simulation tools\cite{todorov2012mujoco, wu2022optimizing, brockman2016openai}. For instance, in the work by \cite{devries2017dataset}, a task-agnostic feature augmentation approach is proposed to enrich training data with minimal additional computation. The authors of \cite{kumar2019closer} also explore similar ideas in the domain of few-shot learning. Building upon this, the concept is adapted to sentence representation learning in works like \cite{yan2021consert, gao2021simcse}. More recently, \cite{li2022metaug} applies feature augmentation based on a meta-learning technique.

\section{Method}
\label{sec: method}

In this section, we first introduce the overall process of hallucination for contrastive representation learning in Section~\ref{sec: pipline}. Secondly, we highlight the center cropping method we used, which is crucial to effective hallucination or generation of new samples in Section~\ref{sec: centercrop}. Then, we introduce the \textit{Hallucinator} incorporated into our contrastive models in Section~\ref{sec: hallucinator}. Finally, we visualize and discuss the critical properties of the hallucination method from two perspectives in Section~\ref{sec: visualization}: the similarity of positive samples and the uniformity of the feature distribution. 

\subsection{The Overall Pipeline}
\label{sec: pipline}
Taking MoCo \cite{chen2020improved} as an example, we illustrate how a \textit{Hallucinator} can be plugged into a contrastive learning model in Figure~\ref{fig: pipline}. Our architecture takes one image $x$ as input. Then, the input $x$ is augmented into two views $x_{1}$ and $x_{2}$. Each view will be processed by an encoder consisting of a backbone (e.g., ResNet) and a projector (e.g., an MLP head). After the encoders, output vectors $q$ and $k$ are obtained, forming one positive pair $(q, k)$. Then, this positive pair $(q, k)$ is fed to a \textit{Hallucinator}. Notably,  \textit{Hallucinator} is only added to one branch of the framework to generate an additional positive feature $\hat{q}$. Together with feature vector $k$, $\hat{q}$ and $k$ form as an extra positive pair  $(\hat{q}, k)$ during the training. Based on different contrastive learning models, the loss functions keep intact, and the average loss of these two positive pairs is computed for back-propagation. The same paradigm could be applied to SimCLR, SiamSiam and other contrastive learning models. We illustrate further details about pipelines and loss functions of other models in this paper in the Supplementary Material (Section 1.1). 

\subsection{Center Cropping}
\label{sec: centercrop}

In contrastive learning, data augmentations aim to ensure the performance of pre-trained representations invariant to nuisances. Among all these methods, random crop plays the most critical role in all the contrastive learning models. Generally, views (cropped tiles) generated by random cropping are diversified, successfully covering all the semantic information over the whole image. However, such a cropping method is likely to generate false positives patches \cite{peng2022crafting}. In other words, patches randomly cropped from the original images do not necessarily share the overlapped pixels and sufficient common information. Therefore, these false positive pairs may be fooling models during training, causing representations to be sub-optimal. Importantly, the issue will be exacerbated if we generate further hallucinated samples based on false positive pairs, which misleads  the overall training beyond the sweet spot.

To tackle this issue, we first apply center cropping $\mathbb{C}_{crop}$ to the original image, getting a relatively smaller image $\hat{I}_{x, y}$. Then, random cropping $\mathbb{R}_{crop}$ is applied to  $\hat{I}_{x, y}$. More specifically, the center cropping can be formulated as 
\begin{align}
        \hat{I}_{x, y} =  \mathbb{C}_{crop}(I_{x, y},  p),
\end{align}
where $I_{x, y}$ is the input image with $(x, y)$  as the coordinate of the images' center. After center cropping, we keep the center of $\hat{I}_{x, y}$ unchanged. Meanwhile, with the original shape of  $I_{x, y}$ defined as $(h,w)$, we define the shape of cropped image $\hat{I}_{x, y}$ as $(\hat{h},\hat{w})$. The $p$ denotes a ratio of cropped length over the original length, i.e., $p=\frac{\hat{w}}{w} = \frac{\hat{h}}{h}$. Only if particularly mentioned, we set $p=0.5$ for all experiments in this paper. 

While center cropping effectively avoids false positive pairs, it reduces the operable region for random cropping and generates positive views with a similar appearance. We, therefore, adopt center-suppressed sampling \cite{peng2022crafting} with a sampling method following a beta distribution $\beta(\alpha,\alpha)$ (i.e., a U-shaped distribution). Concretely, $\beta(\alpha,\alpha)$ assigns a lower probability to the center of the $\hat{I}_{x, y}$ and gives greater probability to its boundary, increasing the variance between views $x_{1}$ and $x_{2}$. Together, we summarize the process to obtain these two views as
\begin{align}
        &x_{1} =  T(\mathbb{R}_{crop} (\hat{I}_{x,y}|\alpha)), \quad \textrm{s.t.} \quad \alpha < 1 \nonumber \\ 
        &x_{2} =  T(\mathbb{R}_{crop} ({I}_{x,y}|\alpha)), \quad \textrm{s.t.} \quad \alpha < 1  ,
\end{align}
where $T$ denotes data augmentations, including color jittering, random grayscale, Gaussian blur and horizontal flipping.  $\mathbb{R}_{crop} (...|\alpha)$ represents random cropping following the $\beta(\alpha,\alpha)$ distribution. $\alpha$ is set to less than 1 to ensure an increasing sampling probability as the pixel's coordinates go beyond the center. A visualization of the center sampling method can be found in Supplementary Material (Section 1.2). 

\begin{figure}[t!]
\begin{center}
    \includegraphics[width=1.05\linewidth]{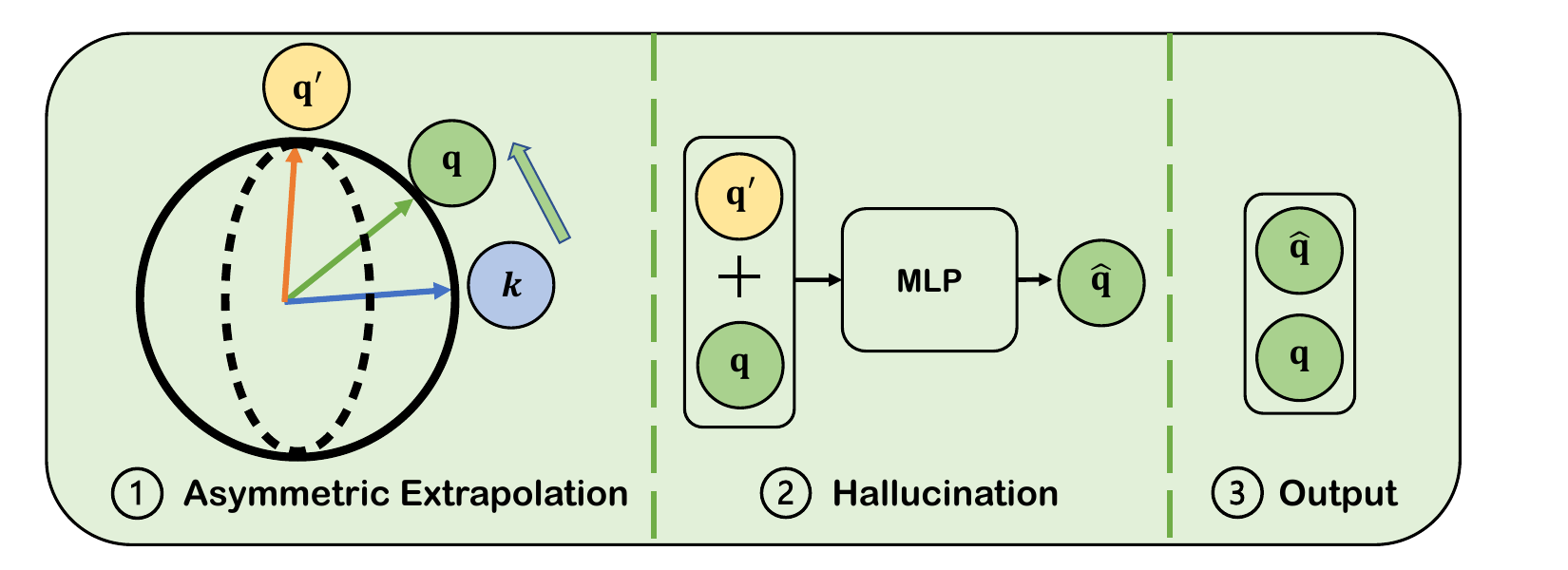}
\end{center}
\vspace*{-3mm} 
  \caption{The $Hallucinator$. Stage 1: The original feature vector $q$ is extrapolated to the opposite direction of feature vector $k$, forming $\hat{q}$ in a linear way. Stage 2: We introduce non-linear transformation to smooth extrapolated features concatenated with $q$ and $q'$. Stage 3: output original $q$ and hallucinated $\hat{q}$.}
  \vspace*{-3mm} 
\label{fig: Hallucinator}
\end{figure}

\subsection{Hallucinator}
\label{sec: hallucinator}
\noindent \textbf{Asymmetric Feature Extrapolation.}
The first objective of this module is to introduce an additional positive pair without introducing extra computations. Therefore, the feature-level operation is preferred compared to image-level operations. To achieve this, \textit{Hallucinator} is plugged in after the views $x_{1}$ and $x_{2}$ are encoded. As a result, the hallucinated (generated) feature is purely based on the two feature vectors $q$ and $k$. 

Meanwhile, since harder positives improve pre-trained encoders' generalization capacity \cite{tian2020makes}, the hallucinated features in positive pairs are favorable if they share less mutual information. Previous work illustrates that a symmetric positive extrapolation is effective in generating hard examples for MoCo \cite{zhu2021improving}. Concretely, two positive features are combined with weighted addition, pushing positive features apart. However, the hallucination process is asymmetric, i.e., \textit{Hallucinator} is only added in one of the branches of the model. Then, we propose to apply the singe-side feature extrapolation to the feature vector $q$, as shown in the first stage of Figure~\ref{fig: Hallucinator}. Additionally, we simplify the sampling strategy of weights for extrapolation from a beta distribution to a uniform distribution. To be specific, we summarize the positive extrapolation in our method as follows:
\begin{align}
\label{eq: extrapolation}
        &{{q}}^{\prime} = (1 + \lambda)  {q} - \lambda {k}   \quad   \textrm{s.t.} \quad \lambda \sim  U(\beta_{1}, \beta_{2}),
\end{align}
where $\lambda$ is sampled from a uniform distribution $U(\beta_{1}, \beta_{2})$. $\beta_{1}$ and $\beta_{2}$, which are the boundary of the uniform distribution, are set to 0 and 0.1 by default. 

\noindent \textbf{Hallucination.}
Positive extrapolation is based on mixup \cite{zhang2017mixup}, i.e., a linear transformation. While positive extrapolation has been proven beneficial to generating hard examples, this linear feature transformation might have a relatively limited capacity to synthesize new feature vectors. This assumption is based on the more satisfactory performance of the non-linear mixup over the original one \cite{guo2020nonlinear}. Similarly, if non-linearity is introduced to the feature generation, the generated vector will benefit more from the training and boost the performance of downstream tasks.

To empower our model with the capacity of non-linear fitting, we introduce the hallucination process. Specifically, we first concatenate $q'$ from the equation~\ref{eq: extrapolation} and the feature vector $q$ together. Then, we use the concatenated feature $(q,q')$ as an input of a non-linear transformation function $H_{\theta}(.)$ that can be instantiated with $n$ linear layers and a ReLU layer between two successive layers. When $n=0$,  $H_{\theta}(.)$ is a non-parametric module, forming an identity function. Such a setting performance is relatively sub-optimal, as shown in the ablation study in Table~\ref{ablation: n-value}. Empirically, we find that  $n= \left\{ 2, 3\right\}$ performs well as the $\textit{hallucinator}$ becomes non-linear and more powerful. We set  $n= 3$ by default as its results are slightly better. With the transformation function $H_{\theta}(.)$, the hallucinated feature is defined as 
\begin{align}
\label{eq: hallucination}
       {\hat{q}}=H_{\theta}({q},{q}^{\prime}) ,
\end{align}
where $\theta$ is the parameters of $H_{\theta}(.)$. Notably,  $H_{\theta}(.)$ is differentiable, allowing us to back-propagate the loss of contrastive learning. Therefore, we update not just the parameters of the encoders and projectors but also the parameters $\theta$ of the hallucinator. The second stage of Figure ~\ref{fig: Hallucinator} illustrates the proposed hallucination process. Following that, we take $q$ and $\hat{q}$ as the output of \textit{Hallucinator}. 




\begin{figure}[t!]
\begin{center}
    \includegraphics[width=0.9\linewidth]{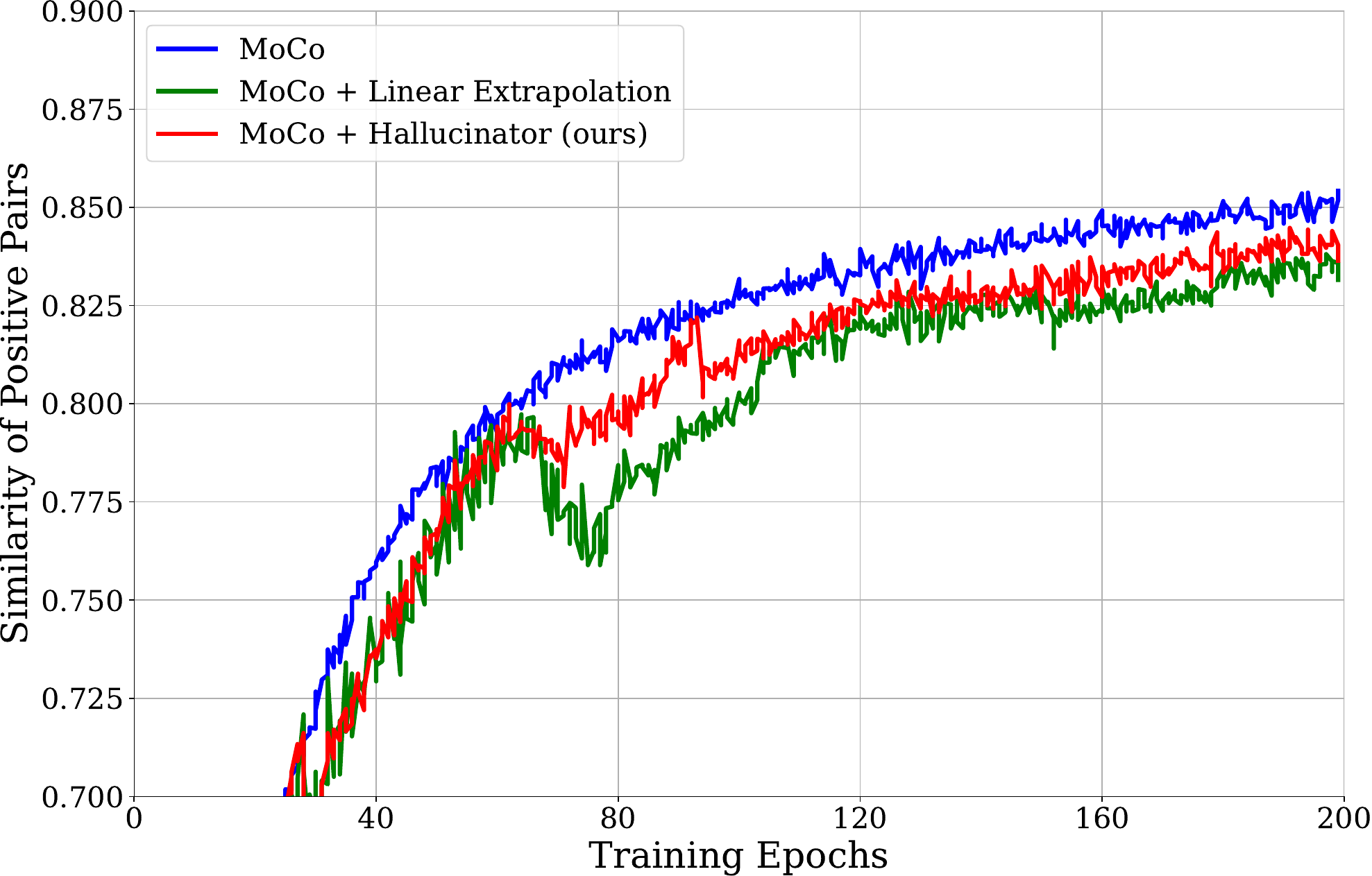}
\end{center}
\vspace*{-5mm} 
  \caption{Similarity of positive pairs in training. Smaller values indicate less mutual information and better representation \cite{tian2020makes,peng2022crafting,zhu2021improving}. As \textit{Hallucinator} incorporates non-linearity extrapolation, it guarantees smooth training and harder positive features.}
  \vspace*{-3mm} 
\label{fig: similarity}
\end{figure}

\subsection{Discussion and Visualization}
\label{sec: visualization}
To better understand the behavior of \textit{Hallucinator}, we discuss two critical properties that may contribute to and explain its effectiveness. For visualization, we train MoCoV2 \cite{chen2020improved} with a standard ResNet-18 \cite{he2016deep} on Cifar-10 \cite{krizhevsky2009learning}.

\noindent \textbf{Similarity of Positive Pair.}
We first investigate the similarity of positive pairs during the training process. Concretely, we quantize mutual information using the cosine similarity of positive pairs ${S}$:
\begin{align}
    {S} \triangleq \frac{q \cdot k}{\left\| {q} \right\| \left\| {k} \right\|} .
\end{align}
While hard positives share less mutual information, thus having a smaller cosine similarity value, the performance of downstream tasks will be enhanced \cite{zhu2021improving}. Based on Figure~\ref{fig: similarity}, the proposed \textit{Hallucinator} generates harder positives with smaller values of similarity. Consequently, it helps contrastive learning models obtain more nuisances-invariant features. More importantly, different from linear symmetric extrapolation, our training curve is relatively stable without many oscillations introduced. This observation indicates that \textit{Hallucinator} is successfully optimized with the overall framework. Meanwhile, hallucinated features nicely fit into the contrastive learning task. 

\begin{figure*}[t!]
\begin{center}
    \includegraphics[width=0.9\linewidth]{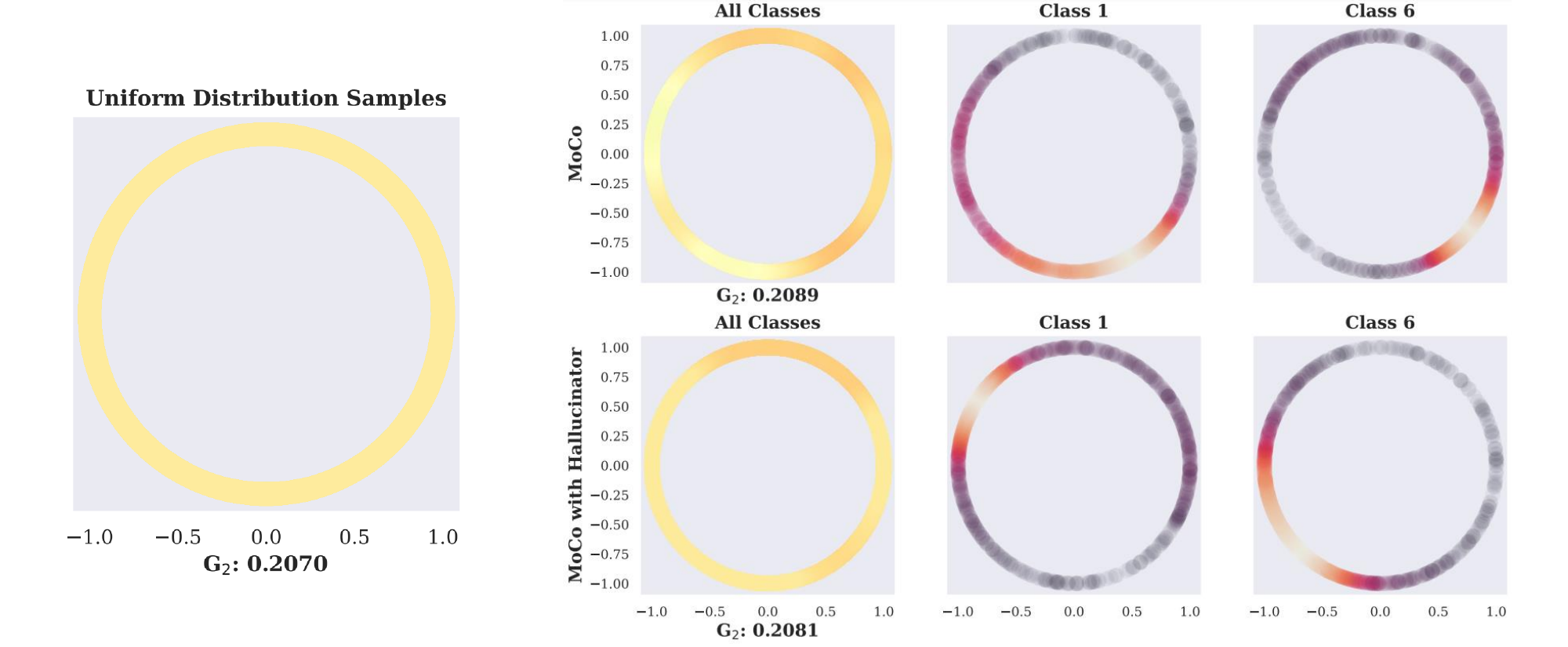}
\end{center}
\vspace*{-3mm} 
  \caption{A measure of uniformity based on $G_{2}$ potential. We plot 10,000 feature vectors with Gaussian kernel density estimation (KDE) in $\mathbb{R}^{2}$. The left subplot illustrates the feature vectors from a uniform distribution. The three feature distributions on the right in the first row visualize the features from MoCoV2 \cite{chen2020improved}. The other three feature distributions in the second row demonstrate MoCoV2 with $\textit{Hallucinator}$. $\textit{Hallucinator}$ benefits the uniformity with a smaller value of $G_{2}$.} 
  \vspace*{-3mm} 
\label{fig: Uniformity}
\end{figure*}

\begin{table*}[t!]
    \centering
    \resizebox{1.5 \columnwidth}{!}{
    \begin{tabular}{ l  ccccccccccc}
    \hline
    \toprule
    \textbf{Dataset}  &\multicolumn{2}{c}{\textbf{CIFAR-10}}  &\multicolumn{2}{c}{\textbf{CIFAR-100}} &\multicolumn{2}{c}{\textbf{Tiny ImageNet}} &\multicolumn{2}{c}{\textbf{STL-10}} \\
    \midrule
    \textbf{Hallucinator} & \xmark & \cmark & \xmark & \cmark & \xmark & \cmark & \xmark & \cmark  \\
    \midrule
    MoCoV1  & 88.31 & \textbf{88.94} & 60.94 & \textbf{61.81} & 44.65 & \textbf{45.53} & 88.19 & \textbf{90.09}\\
    MoCoV2  & 87.21 & \textbf{89.23} & 59.70 & \textbf{61.26} & 47.12 & \textbf{47.95} & 89.32 & \textbf{90.46}\\
    SimCLR  & 89.66 & \textbf{90.11} & 60.94 & \textbf{61.43} & 45.22 & \textbf{46.30} & 89.07 & \textbf{89.98}\\
    SimSiam & 90.47 & \textbf{90.78} & 63.39 & \textbf{64.38} & 43.66 & \textbf{44.96} & 87.79 & \textbf{88.16}\\
    \bottomrule

    \end{tabular}
    }
    
    \caption{
    Linear classification results for different contrastive methods and datasets in small scales. We adopt ResNet-18 as the backbone and report the classification results with or without $\textit{Hallucinator}$.
    }
    \label{tab:classification-small}
\end{table*}

\noindent \textbf{Uniformity.}
We continue to analyze the performance of the proposed model from the perspective of uniformity. Notably, feature vectors from contrastive learning should be roughly uniformly distributed on a unit hyper-sphere, which ensures maximal information is preserved in the feature space \cite{wang2020understanding}.  In other words, the closer the feature distribution is to the uniform distribution, the more the feature benefits downstream tasks. To quantize uniformity, we follow previous work \cite{wang2020understanding} to compute the average value of the Gaussian potential kernel (i.e., Radial Basis Function kernel) of positive features:

\begin{align}
    G_{t}(q,k) \triangleq e^{-t \left\|q - k \right\|^{2}_{2}}  \quad   \textrm{s.t.} \quad t > 0,
\end{align}
where $t$ is a fixed parameter set as 2 for all the experiments. We visualize the feature vectors by mapping them to two-dimension feature space and applying $l_{2}$ normalization in Figure~\ref{fig: Uniformity}. $G_{2}=0.2070$ for uniformly distributed samples, whereas features from MoCoV2 have $G_{2}=0.2089$. If we plugin \textit{Hallucinator} into MoCo, \textit{Hallucinator} provides further contrast during the training with extra positives introduced, giving more uniformly distributed features and a decreased $G_{2}$ value,i.e., 0.2081. Additionally, we visualize the features of two classes of Cifar-10. Each of these features is well-clustered. With better uniformity, clusters' overlapping decreases, forming more linearly separable features. Therefore, we could observe that the overlapping of feature clusters between class 1 and class 6 in MoCo is larger than the one with \textit{Hallucinator} plugged in.

\section{Experiments and Results}
In this section, we conduct various experiments on different datasets and contrastive learning models to demonstrate the effectiveness of the \textit{Hallucinator}. Firstly, we introduce the datasets and details in experiments in Section~\ref{sec: data and train}. Then, we continue to evaluate the performance of the proposed method using linear probing protocol following the paradigm of previous work \cite{chen2020simple,chen2020improved,chen2021exploring, grill2020bootstrap} in Section~\ref{sec: linear classification}. Following this,  we conduct ablation studies to understand how each module contributes to the final results in Section~\ref{sec: ablation}. Lastly, we show the transferability of pre-trained encoders in downstream tasks requiring dense pixel predictions, including object detection and semantic segmentation in Section~\ref{sec: transfer}. 

\subsection{Datasets and Training Details}
\label{sec: data and train}
\noindent \textbf{Datasets and Baseline Models.} 
We first evaluate the performance of \textit{Hallucinator} based on a wide range of datasets crossing different scales. Specifically, these datasets include CIFAR-10, CIFAR-R100 \cite{krizhevsky2009learning}, Tiny ImageNet, STL-10 \cite{coates2011analysis} and ImageNet \cite{deng2009imagenet}. Meanwhile, we demonstrate the performance of \textit{Hallucinator} in several popular contrastive learning frameworks, including MoCoV1 \cite{he2020momentum}, MoCoV2 \cite{chen2020improved}, SimCLR \cite{chen2020simple} and SimSiam \cite{chen2021exploring}. 

\noindent \textbf{Training and Evaluation Details.} 
Importantly, we strictly adopt the same training settings when evaluating the \textit{Hallucinator}'s performance. While additional gain could be achieved with further exploration of hyper-parameter, it is not the focus of this work. The ultimate goal of the proposed \textit{Hallucinator} is to provide further feature-level contrast for self-supervised learning. With the \textit{Hallucinator} plugin, it benefits diversified self-supervised learning methods regardless of their type of backbones and hyper-parameters of training. 

Datasets on relatively small scales include CIFAR-10\&100, Tiny ImageNet and STL10. To ensure fair comparisons, we keep training settings the same over all these datasets. We follow the paradigm of previous work for pre-training \cite{peng2022crafting}. Concretely, we pre-train contrastive learning models for 500 epochs with a batch size of 512 and ResNet-18 as the backbone. To optimize the models, we use an SGD optimizer and a cosine-annealed learning rate of 0.5. In the linear classification task, we train models for 100 epochs. With the initial learning set as 10, we divide it by 10 at the 60th and 80th epochs.

As the scale of the dataset expands to ImageNet, we use ResNet-50 as the backbone. For MoCoV1, MoCoV2 and SimSiam, all the pre-training settings follow the original work. We use a batch size of 512 and an SGD optimizer with a cosine-annealed learning rate of 0.5 for SimCLR, the same as the settings in \cite{peng2022crafting}. For the linear classification task, we follow the original work in \cite{he2020momentum}, training the model for 100 epochs. We set the initial learning rate to 30. Then, we divide the learning by 10 at the 60th and 80th with a weight decay of 0. 

We set $p=0.5$ and $\alpha=0.6$ for center cropping and center-suppressed sampling, respectively. For asymmetric linear extrapolation, we set $\beta_{1}=0$ and $\beta_{2}=0.1$. In the Hallucination process, $n = 3$ for all the results reported in this paper. All the experiments are conducted on a server with 8 GPUs.

\begin{table}
    \centering
    \resizebox{0.95 \columnwidth}{!}{
    \begin{tabular}{ l | cccc}
    \hline
    \toprule
    \textbf{Method}  &\textbf{Backbone}  &\textbf{Epoch} &\textbf{IN-200} &\textbf{IN-1K} \\

    \midrule
    MoCoV1        & ResNet-50 &100 & 62.19 & 57.27 \\
    MoCoV1(Ours)  & ResNet-50 &100 & \textbf{63.46} & \textbf{59.17} \\
    \midrule
    MoCoV2        & ResNet-50 &100 & 62.57 & 64.41 \\
    MoCoV2(Ours)  & ResNet-50 &100 & \textbf{63.58} & \textbf{64.97} \\    
    \midrule
    SimCLR        & ResNet-50 &100 & 62.22 & 61.23 \\
    SimCLR(Ours)  & ResNet-50 &100 & \textbf{63.02} & \textbf{61.71} \\
    \midrule
    SimSiam        & ResNet-50 &100 & 62.80 & 63.11 \\
    SimSiam(Ours)  & ResNet-50 &100 & \textbf{63.52} & \textbf{63.55} \\

    \bottomrule

    \end{tabular}
    }
    \caption{
    Linear classification results on IN-200 and IN-1K. Our method plugged in the proposed \textit{Hallucinator} in baseline methods. All the models are pre-trained for 100 epochs and use identical training settings for fair comparisons.  
    }
    \label{tab:classification-big}
\end{table}

\subsection{Linear Classification Protocol}
\label{sec: linear classification}
Following the previous protocol, we first evaluate the proposed method by linear classification of frozen features. For each dataset, we report the top-1 classification accuracy on the validation set.

\noindent \textbf{Results on Small-Scale Datasets.}
The classification results on Cifar10\&100, Tiny ImageNet and STL-10 are reported in Table ~\ref{tab:classification-small}. With \textit{Hallucinator} introduced, we notice a stable improvement over the baselines ranging from $0.31\%$ to $1.98\%$. Notably, such improvements do not introduce extra computations and generalize well to various models.

\noindent \textbf{Results on ImageNet.}
For the results of ImageNet, we report the results at two different scales. First, we evaluate its performance on standard IN-1K (i.e. ImageNet-1K), which consists of 1000 classes. Second, we test the proposed method in IN-200 (i.e. ImageNet-200) with 200 randomly selected classes. We report the corresponding results in Table~\ref{tab:classification-big}. We found that $\textit{Hallucinator}$ essentially benefits MoCoV1 with $1.27\%$ and $1.89\%$ improvements for IN-200 and IN-1K accordingly. For MoCoV2, SimSiam and SimCLR, we notice a gain in accuracy ranging from  $0.44\%$ to  $1.01\%$. On average, the gains are more salient in IN-200.

\begin{table}
    \centering
    \resizebox{0.93 \columnwidth}{!}{
    \begin{tabular}{ l | ccc}
    \hline
    \toprule
    \textbf{Acc.(\%)}  &\textbf{\makecell[c]{Center\\ Cropping }}  &\textbf{\makecell[c]{Asymmetric\\ Extrapolation }} &\textbf{Hallucination}  \\

    \midrule
    62.57        &\xmark  &\xmark &\xmark  \\
    63.58(+1.01)        & \cmark &\cmark & \cmark \\
    \midrule
    62.58(+0.01)        & \cmark &       &   \\   
    62.82(+0.25)        &        & \cmark      &   \\   
    62.84(+0.27)        &        &       &\cmark   \\  
    63.07(+0.50)        &        &\cmark       &\cmark   \\  

    \bottomrule

    \end{tabular}
    }
    \caption{
    Ablation of the different modules in the proposed method.
    }
    \label{ablation: modules}
\end{table}

\begin{table}
    \centering
    \resizebox{0.95 \columnwidth}{!}{
    \begin{tabular}{ l | cccccc}
    \hline
    \toprule
    \textbf{p (\%)}  & {90}  & {80} & {70} & {60} & {50} &{40} \\

    \midrule
    \textbf{Acc.(\%)}  &63.11 &63.19 &63.33 &63.56 &63.58 &63.21   \\

    \bottomrule

    \end{tabular}
    }
    \caption{
    Ablation of classification results w.r.t the $p$ value.
    }
    \label{ablation: p-value}
\end{table}

\begin{table}
    \centering
    \resizebox{0.97 \columnwidth}{!}{
    \begin{tabular}{ l | ccccc}
    \hline
    \toprule
    \textbf{(\bm{$\beta_{1}, \beta_{2}$})}  & \textbf{(-0.5, 0.0)}& \textbf{(0.0, 0.5)} &  \textbf{(0.5, 1.0)}  & \textbf{(1.0, 1.5)} & \textbf{(0.0, 1.0)}  \\

    \midrule
    \textbf{Acc.(\%)}  &62.80 &63.32 &63.55 &63.42 &63.58  \\

    \bottomrule

    \end{tabular}
    }
    \caption{
    Ablation of classification results w.r.t the extrapolation range.
    }
    \label{ablation: ranges}
\end{table}

\begin{table}
    \centering
    \resizebox{0.9 \columnwidth}{!}{
    \begin{tabular}{ l | ccccc}
    \hline
    \toprule
    \textbf{layer \#: n}  & {0}  & {1} & {2} & {3} &{4}  \\

    \midrule
    \textbf{Acc.(\%)}  &62.85 &63.03 &63.57 &63.58 &63.55   \\

    \bottomrule

    \end{tabular}
    }
    \caption{
    Ablation of classification results w.r.t the number of linear layers $n$.
    }
    \label{ablation: n-value}
\end{table}

\begin{table*}
    \centering
    \resizebox{2.1 \columnwidth}{!}{
    \begin{tabular}{ l | c | ccc | ccc | ccc }
    \hline
    \toprule
    \textbf{Method}  &\multicolumn{1}{c|}{\textbf{1N-1k}}  &\multicolumn{3}{c|}{\textbf{VOC detection}} &\multicolumn{3}{c|}{\textbf{COCO detection}} &\multicolumn{3}{c}{\textbf{COCO instance segmentation}}  \\
                  & Top-1    &$AP$ & $AP_{50}$ &$AP_{75}$ &$AP^{bb}$ & $AP^{bb}_{50}$ &$AP^{bb}_{75}$  & $AP^{mk}$ &$AP^{mk}_{50}$ &$AP^{mk}_{75}$     \\
    \midrule
    Random init  & -    &33.8 & 60.2 & 57.27  &26.4  &44.0  &27.8    &29.3  &46.9  &30.8   \\
    Supervised   & 76.1 &53.5 &81.3 &58.8 &38.2 &58.2 &41.2  &33.3 &54.7 &35.2 \\
    InfoMin \cite{tian2020makes}   & 70.1 &57.6 &82.7 &64.6 &39.0 & 58.5 & 42.0  &34.1 & 55.2 &36.3 \\
    \midrule
    MoCoV1 \cite{he2020momentum}        & 60.6 &55.9 &81.5 &62.6 &38.5 &58.3 &41.6  &33.6 &54.8 &35.6 \\
    MoCoV1(Ours)  & \textbf{63.8} &\textbf{56.7} & \textbf{81.8} & \textbf{63.2} &\textbf{38.9} & \textbf{58.5} & \textbf{41.9} &\textbf{33.8} & \textbf{55.2} & \textbf{36.0}  \\    
    \midrule
    MoCoV2 \cite{chen2020improved}        & 67.5 &57.0 &82.4 &63.6 &39.0 &58.6 &41.9  &34.2 &55.4 &36.2 \\
    MoCoV2(Ours)  & \textbf{68.0} &\textbf{57.4} &\textbf{82.7} &\textbf{63.9} &\textbf{39.3} &\textbf{58.8} &\textbf{42.3} &\textbf{34.6} &\textbf{55.5} &\textbf{36.4} \\    

    \bottomrule

    \end{tabular}
    }
    \caption{
    Fine-tuning results on object detection tasks on PASCAL VOC and COCO, and instance segmentation on COCO. All models are pre-trained for 200 epochs on ImageNet-1K.
    }
    \label{tab: transfer learning}
\end{table*}

\subsection{Ablation Studies}
\label{sec: ablation}
In this section of ablation studies, we investigate essential modules in our proposed methods, including center cropping, asymmetric extrapolation and the process of hallucination. We conduct experiments using MoCoV2 with ResNet-50 as the backbone. We report the results of the classification task in IN-200 for all the following experiments.

\noindent \textbf{Contributions of Modules.}
We first report the results with or without crucial modules introduced in Section ~\ref{sec: method}. According to Table~\ref{ablation: modules}, center cropping shows a similar performance to the original cropping method, successfully covering major semantic information of images. However, it successfully avoids false positives in pre-training, which is critical for hallucination. Asymmetric extrapolation benefits the performance of representation learning with reduced mutual information. This observation is consistent with the results shown in the symmetric extrapolation \cite{zhu2021improving}. If we combine asymmetric extrapolation and the hallucination method, the performance of the model could be further boosted. However, it is still sub-optimal because of possible false positives in cropping.

\noindent \textbf{Center Cropping.}
In this work, we apply center cropping as a critical tool in one branch of contrastive learning to avoid false positives. Therefore, we investigate how the proposed model is influenced by the length ratio $p$ in Table~\ref{ablation: p-value}. The classification results are favorable when p is in the range (0.5, 0.6). We notice a performance drop when p is less than 0.5. This drop is because the center cropping failed to cover all semantic information in the images.

\noindent \textbf{Extrapolation Range.}
We continue to report the extrapolation range given its differences in sampling method and single-side architecture compared with the previous method \cite{zhu2021improving}. Based on Table~\ref{ablation: ranges}, the classification accuracy drops if we set $\beta_{1}=-0.5$ and  $\beta_{2}=0$, respectively. Such observation is because hallucinated positives share more mutual information, thus less beneficial to training. Generally, extrapolation boosts accuracy by outperforming the baseline by at least 0.75$\%$. However, the improvement decreases when $\beta_{1}>1.0$. We eventually set $\beta_{1}=0.0$ and $\beta_{2}=1.0$, given its best performance.

\noindent \textbf{Hallucination.}
We investigate how the number of linear layers influences the model's performance. When $n=0$, the hallucination process is an identity transformation, introducing no additional operation except for extrapolation. When $n=2$, non-linearity is introduced. Then we notice a salient improvement over the baseline. We set $n=3$ by default since it gives better performance.

\subsection{Transferring Features}
\label{sec: transfer}
The primary goal of representation is to learn transferrable features. We evaluate the transferability of features from the proposed method following the previous protocol \cite{he2020momentum,chen2020improved,chen2020simple,chen2021exploring}. Then, we compare the representation quality by transferring them to downstream tasks, including VOC \cite{everingham2009pascal} object detection and COCO \cite{lin2014microsoft} object detection and instance segmentation. Notably, we re-implement all these experiments using the same settings in MoCo's detectron2 codebase \cite{wu2019detectron2}.

\noindent \textbf{Object Detection on PASCAL VOC.}
Following the paradigm of previous work \cite{he2020momentum}, we use Faster R-CNN\cite {girshick2015fast} as the object detection method using R50-C4 as the detector \cite{he2017mask}. We train the model end-to-end on the \textbf{trainval2007+2012} and evaluate its performance on \textbf{test2007}. As shown in Table~\ref{tab: transfer learning}, we observe a stable gain range from 0.3 to 0.8 under different metrics on MoCoV1 and MoCoV2. 

\noindent \textbf{Object Detection and instance segmentation on COCO.}
We continue to report the detection and segmentation results on COCO using  Mask R-CNN  \cite{he2017mask}. Similarly, we use the R50-C4 as the backbone. The is model in an end-to-end way on \textbf{train2017}. Then, the model is evaluated on \textbf{val2017}. Again, the proposed method benefits the object detection and segmentation tasks with various metrics as demonstrated in Table~\ref{tab: transfer learning}. 



\section{Conclusion}
In this work, we propose \textit{Hallucinator}, which generates additional hard positive pairs for contrastive learning models based on Siamese structure. \textit{Hallucinator} generates novel data samples in the feature space to provide the training with further contrast without additional computation. We design an asymmetric feature extrapolation to avoid trivial positive pairs and innovatively introduce non-linear hallucination to smooth the generated samples. We empirically prove the effectiveness and generalization capacity of \textit{Hallucinator} to well-recognized contrastive learning models, including MoCoV1\&V2, SimCLR and SimSiam. Finally, we hope this work could bring the concept of ``Hallucination'' into the SSL domain and unlock future research on sample generations\&synthesis in contrastive learning.

{\small
\bibliographystyle{ieee_fullname}
\bibliography{egbib}
}

\end{document}